# Generative Modeling in Structural-Hankel Domain for Color Image Inpainting

Zihao Li, Chunhua Wu, Shenglin Wu, Wenbo Wan, Yuhao Wang, *Senior Member, IEEE*,
Qiegen Liu, *Senior Member, IEEE*

*Abstract*—In recent years, some researchers focused on using a single image to obtain a large number of samples through multi-scale features. This study intends to a brand-new idea that requires only ten or even fewer samples to construct the low-rank structural-Hankel matrices-assisted score-based generative model (SHGM) for color image inpainting task. During the prior learning process, a certain amount of internal-middle patches are firstly extracted from several images and then the structural-Hankel matrices are constructed from these patches. To better apply the score-based generative model to learn the internal statistical distribution within patches, the large-scale Hankel matrices are finally folded into the higher dimensional tensors for prior learning. During the iterative inpainting process, SHGM views the inpainting problem as a conditional generation procedure in low-rank environment. As a result, the intermediate restored image is acquired by alternatively performing the stochastic differential equation solver, alternating direction method of multipliers, and data consistency steps. Experimental results demonstrated the remarkable performance and diversity of SHGM.

## I. INTRODUCTION

Image inpainting is the process of completing or recovering the missing region in the image or removing some objects added to it [1]-[7]. With the increasing interest in image inpainting, many approaches have emerged, which can be broadly classified into three categories: Diffusion-based methods, patch-based methods, and deep learning-based methods. Specifically, the kind of diffusion-based image inpainting [8]-[11] smoothly spreads the image content from the boundary to the interior of the missing region in the image and gradually synthesizes new textures to fill these areas. In addition, patch-based methods [12]-[16] search for the most matching similar patch in the visible area according to the internal distribution within images. Especially, generative models, as one of the deep learning-based methods [17]-[23], had achieved remarkable successes in image inpainting. The representative pioneer approaches included generative adversarial networks (GANs) [24]-[27] and score-based generative models [28]-[30].

On one hand, GANs become the most used technique in image inpainting due to its great potential in fitting data distributions with adversarial mechanisms. For example, Yu *et al.* [31] considered an encoder-decoder backbone to design the coarse and refinement networks as a two-stage network and used the contextual attention module of Iizuka *et al.* [32] to modify their network. Brock *et al.* [19] trained GAN at the largest scale yet attempted and focused on the instabilities specific to such scale. Undoubtedly, the model demonstrably boosted performance at the expense of consuming a long time. Due to the lack of a great deal of training data, GANs are prone to over-fitting and often lead to mode collapse [20] as well as training instabilities [21]. On the other hand, score-based generative models [28]-[30] have gained remarkable success as a novel class of generative models, which can produce high-quality image samples comparable to GANs, without adversarial optimization. For that, Song *et al.* [29] provided a new theoretical analysis of learning and sampling from score-based models in high-dimensional spaces, explaining existing failure modes and motivating new solutions for generalization across datasets.

Due to infeasible or insufficient data for some studies, a few researchers have turned to learning internal statistics or modeling the internal distribution of patches within a single natural image. For instance, SinGAN, designed by Shaham *et al.* [20], is an unconditional generative scheme learned from a single natural image and is not limited to texture images. This allows generating new samples of arbitrary size and aspect ratio. Besides, Sushko *et al.* [22] presented an unconditional generative model One-Shot GAN with a two-branch discriminator that merely operating at a training dataset with a single image. What's more, Zheng *et al.* [23] took efficiency into consideration by combining an energy-based generative framework with a coarse-to-fine sequential training and sampling strategy. To capture internal statistical properties at different scales, large images are successively subsampled for attaining sufficient small image patches as the training set in the above references. In summary, the common denominator of these efforts is exploiting the multi-scale property of images.

Unlike the widespread usability of the multi-scale property to attain sufficient samples for training, we propose a brand-new strategy to achieve the goal. More precisely, several images are employed to construct structural-Hankel matrices based tensors as training samples of the score-based model. These samples with the low-rank and structural properties easily enables the score-based model to better capture the internal statistical distributions within images. This is achieved by two key ingredients: The low-rank and structural properties in the structural-Hankel matrix and the score-based model [28]-[30].

Low-rank matrix completion is an active research area that shares many similarities with compressed sensing theo-

This work was supported in part by National Natural Science Foundation of China under 61871206. (Z. Li and C. Wu are co-first authors.)
Z. Li, C. Wu, S. Wu, W. Wan, Y. Wang, and Q. Liu are with School of Information Engineering, Nanchang University, Nanchang 330031, China. ({lizihao, wuchunhua, wushenglin}@email.ncu.edu.cn, {wanwenbo, wangyuhao, liuqiegen}@ncu.edu.cn)

ry [33], [34]. As a promising candidate for image inpainting, the low-rank structural-Hankel matrix has attracted increasing interest in natural image restoration over the past decades. In particular, the missing pixels in an image can be filled by employing a low-rank structural-Hankel matrix completion approach. To put it differently, the value of each pixel in an image is assumed to be a linear combination of the neighbor pixels and the missing values could be estimated by minimizing a Hankel matrix. For example, Jin *et al.* [35] introduced a patch-based image inpainting method called annihilating filter-based low-rank structural-Hankel matrix (ALOHA), which took advantage of a low-rank structural-Hankel matrix and annihilation property between a shift-invariant filter and image data. Moreover, to address the problems of texture modeling and inpainting, Sznaier *et al.* [36] developed a Hankel operator, which can be found by factoring a Hankel matrix constructed from the image data. Meanwhile, Sasaki *et al.* [37] devoted a new image inpainting algorithm based on the matrix rank minimization and locally linear embedding method. Razavikia *et al.* [38] proposed a method called EMAC that employed the Hankel transform [39] to represent the FRI signal in a low-rank structure, *i.e.*, formulated the problem of reconstructing binary shape images from a few blurred samples as the recovery of a rank $r$ matrix that was formed by a Hankel structure on the pixels. By extending the idea of estimating missing pixels from annihilating filters and the associated rank-deficient Hankel matrices, Jin *et al.* [40] presented a novel impulse noise removal algorithm. It is worth noting that a series of works utilized the structural-Hankel matrix as the starting point, and what they have in common is to make full use of the low-rank property in Hankel matrix.

In this work, the proposed SHGM jointly explores the low-rank and structural properties in the Hankel matrix via robust generative modeling for flexible image inpainting. Unlike the supervised learning method that usually requires thousands or even tens of thousands of samples, this study merely leverages ten or even fewer images to train the scored-based model. More precisely, several images are divided into many internal-middle patches and then construct them into structural-Hankel matrices. Whereafter, the structural-Hankel matrices are folded into higher dimensional tensors for prior learning. Furthermore, by taking advantage of that generative models to learn the distribution of data, the structural-Hankel matrix explicitly represents the internal statistical properties within images. In view of this characteristic, this paper extends the score-based model to the structural-Hankel domain for image inpainting, and infers missing information by estimating the corresponding data density in the structural-Hankel domain through matching network. During the iterative inpainting stage, we employ predictor-corrector (PC) sampler as a numerical stochastic differential equation (SDE) solver and combine the alternating direction method of multipliers (ADMM) with data consistency (DC) operation to achieve higher quality image restoration. In summary, the main contributions of this work are as follows:

- ***Generative Modeling on Tensors from Structural-Hankel Matrices:*** To alleviate the insufficiency of valid data samples in prior information learning, we model tensors from structural-Hankel matrices. Firstly, many internal-middle patches are obtained by extracting patch-by-patch from several images. Secondly, these internal-middle patches are constructed into structural-Hankel matrices. Finally, these structural-Hankel matrices are folded into tensors to enable efficient training of the score-based model.
- ***Conditional Generation for Robust Restoration:*** To faithfully explore the prior knowledge learned by generative modeling from the structural-Hankel domain, we view the image inpainting as a conditional generation in low-rank circumstance. Specifically, the iterative inpainting is implemented by alternating PC sampler, ADMM, and DC to achieve high-quality restoration effect.

The remainder of this paper is presented as follows. Section II briefly reviews some relevant works in this paper. In Section III, we elaborate on the formulation of the proposed method and describe the prior learning and iterative inpainting process of SHGM in detail. Section IV evaluates the inpainting performance of the present method in comparison with state-of-the-arts. Conclusion is given in Section V.

## II. RELATED WORK

### A. Image Inpainting

Mathematically, the formulation of image inpainting problem can be expressed as:
$$y = Dx + e \qquad (1)$$
where $x \in \mathbb{R}^{M_x}$ is the unknown image to be estimated, $y \in \mathbb{R}^{M_x}$ represents the degraded image, $M_x$ denotes the image dimension, and $e$ is additive noise. $D \in \mathbb{R}^{M_x \times M_x}$ denotes the degradation matrix relating to a degraded imaging system.

A classic approach finds the solution to Eq. (1) by solving the following constrained optimization problem:
$$\min_x \|Dx - y\|^2 + \lambda R(x) \qquad (2)$$
where the first and second terms are the DC term and the regularization term, respectively. $\|\cdot\|^2$ represents the $l_2$ norm and $\lambda$ is the factor that balances the DC term and the regularization term.

To solve Eq. (2), various prior knowledges are integrated into the regularization term $R(x)$ to obtain stable and high-quality solutions. For example, a sparsity-promoting regularize derived from compressed sensing theory [41], [42], such as $l_1$ wavelet [42], total variation [43], *etc*. Furthermore, the supervised end-to-end learning approaches essentially use a discriminative fashion to learn an implicit prior, lacking flexibility and robustness. In this work, we turn to the explicit prior construction for image inpainting via the score-based generative model.

### B. Structural-Hankel Matrix

The distinctive highlight of the method in this study is to construct the structural-Hankel matrix from the image. It can be described as combining all the three-channel RGB pixel intensities into a single Hankel matrix, whose columns are vectorized blocks selected by sliding a three-channel window across the entire data. Specifically, by starting with the size of $N_x \times N_y \times 3$ data, a Hankel matrix $A$ of $3w^2 \times (N_x - w + 1)(N_y - w + 1)$ can be generated by sliding a $w \times w \times 3$ window across the entire image data. Fig. 1 demonstrates how to construct the Hankel matrix $A$ by sliding $5\times5\times3$ window.

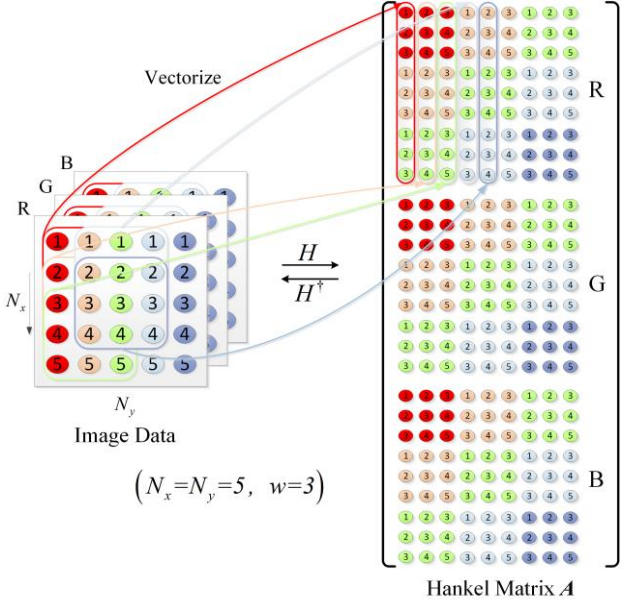

**Fig. 1.** Constructing a Hankel matrix from three-channel R, G, B image data ($H$) and vice versa ($H^{\dagger}$). Individual blocks of data in the image are vectorized as columns in the Hankel matrix.

Due to the property of the sliding window operation, the Hankel matrix $A$ has the block-wise structure, where many pixels from the same image location are repeated in the anti-diagonal directions. As depicted in Fig. 1, the number of matrix rows is heavily larger than that of columns. The linear operator $H$ is defined as the procedure that generates a Hankel matrix $A$ from a three-channel image:

$$H: \mathbb{R}^{N_x \times N_y \times 3} \to \mathbb{R}^{3w^2 \times (N_x-w+1)(N_y-w+1)} \quad (3)$$

Then, the reverse operator $H^{\dagger}$ generates the corresponding image data from the Hankel matrix $A$, and is given as follows:

$$H^{\dagger}: \mathbb{R}^{3w^2 \times (N_x-w+1)(N_y-w+1)} \to \mathbb{R}^{N_x \times N_y \times 3} \quad (4)$$

where $\dagger$ denotes a pseudo-inverse operator. Namely, the operator is equivalent to averaging the anti-diagonal elements and placing them in the appropriate image locations.

During the construction of the structural-Hankel matrix, the same pixel of the image data will exist repeatedly in many different positions of the Hankel matrix. In other words, there is much redundant information in the structural-Hankel matrix, which comprehensively includes the internal statistical properties within the image, as shown in Fig. 2. According to the above construction process, the redundancy phenomenon of structural-Hankel can be formulated as a low-rank matrix optimization:

$$\min_{x} \, rank(A) \quad s.t. \quad x = H^{\dagger}(A) \quad (5)$$

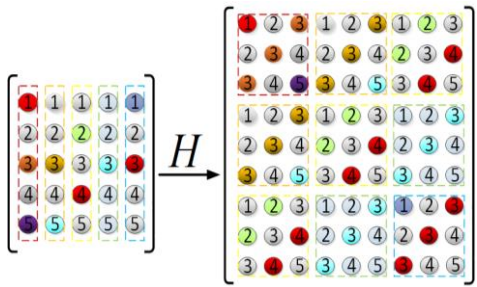

**Fig. 2.** Schematic diagram of the internal statistical property inherent in structural-Hankel matrix.

### C. Score-based Generative Model with SDE

Generative models have achieved great success in generating realistic and diverse data samples [28]-[30]. Among them, score-based matching is a method that estimates the gradient of the log probability density with respect to data by optimizing a score network $s_{\theta}(x)$ parameterized $\theta$. Lately, Song *et al.* [30] presented a score-based generative model with SDE, named NCSN++, which defines a diffusion process (*i.e.*, SDE), as shown in Fig. 3.

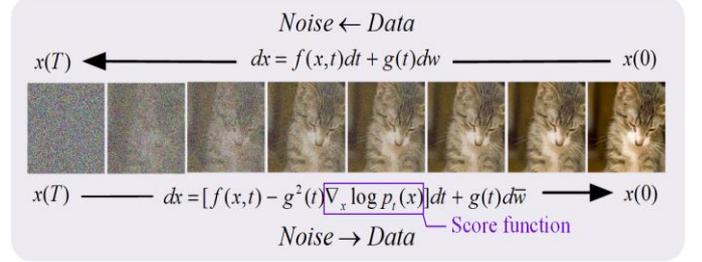

**Fig. 3.** Forward and reverse-time process of SDE. Solving a reverse-time SDE yields a score-based matching generative model. Transforming data to a simple noise distribution can be accomplished with a continuous-time SDE. This SDE can be reversed if we know the score of the distribution at each intermediate time step, i.e., $\nabla_x \log p_t(x)$.

More specifically, a diffusion process $\{x(t)\}_{t=0}^{T}$ with $x(t) \in \mathbb{R}^{M_x}$ is indexed by a continuous-time variable $t \in [0,T]$. There is a dataset of *i.i.d.* samples (*i.e.*, $x(0) \sim p_0$) and a tractable form to generate samples efficiently (*i.e.*, $x(T) \sim p_T$), where $p_0$ and $p_T$ refer to the data distribution and the prior distribution, respectively. Here, the diffusion process can be modeled as the solution of the following SDE:

$$dx = f(x,t)dt + g(t)dw \quad (6)$$

where $f \in \mathbb{R}^{M_x}$ and $g \in \mathbb{R}$ are the drift coefficient and the diffusion coefficient of $x(t)$, respectively. $w \in \mathbb{R}^{M_x}$ is the standard Wiener process.

By starting with samples of $x(T) \sim p_T$ and reversing the process, samples of $x(0) \sim p_0$ can be obtained. The above reverse process is also a diffusion process [45], which can be expressed as the reverse-time SDE:

$$dx = \left[ f(x,t) - g(t)^2 \nabla_x \log p_t(x) \right] dt + g(t) d\overline{w} \quad (7)$$

where $dt$ is the infinitesimal negative time step, $\overline{w}$ is the reverse-time flow process of $w$, and $\nabla_x \log p_t(x)$ is the score of each marginal distribution. Since the true $\nabla_x \log p_t(x)$ of all $t$ is unknown, it can be estimated by training a time-dependent scoring network $s_{\theta}(x_t, t)$, *i.e.*, satisfy $s_{\theta}(x,t) \simeq \nabla_x \log p_t(x)$. Afterwards, samples can be generated by using a numerical SDE solver. Many general-purpose numerical methods exist for solving SDE, such as Euler-Maruyama and stochastic Runge-Kutta methods [44]. Any of them can be applied to the reverse-time SDE for sample generation.

## III. PROPOSED SHGM MODEL

In this section, we introduce the prior learning scheme and implementation details of the proposed SHGM for color image inpainting.

### A. Motivation

The success of deep learning significantly depends on the amount and quality of the training dataset. Particularly, for end-to-end supervised learning, thousands or even tens of thousands of training samples are needed. Although generative approaches have become a popular choice in recent years, they also required hundreds of images for model training. On the aforementioned problem, this work tries to dramatically reduce the training samples to a few given images.

On one hand, inspired by the property of Hankel matrix over the past decades [35]-[40], a structural-Hankel matrix much larger than the original image can be constructed by using algebraic methods. Importantly, due to the structural-Hankel matrix explicitly represents the statistical distribution within an image, the internal statistical property can be easily captured. Based on this characteristic, we only utilize several images to generate Hankel matrices with internal statistical properties as training samples, which are sufficient to effectively learn appropriate priors for image inpainting tasks. In short, several raw image samples are divided into many internal-middle patches, which are then converted into large-scale Hankel matrices and folded into higher-dimensional tensors. In this way, many small-scale structural-Hankel matrix samples with low-rank and structural properties can be obtained. We refer to this process as the construction of structural-Hankel matrices.

On the other hand, we perceive the phenomenon that the score-based model can learn the compositional structure or specific prior distribution of the samples. Consequently, we intend to another direction, that is, the distribution of the structural-Hankel matrices is modeled using a score-based generative model, which learns prior information related to image distribution features. During the iterative inpainting process, it is conditionally sampled by alternatively executing the SDE solver, ADMM, and DC steps. Details of the prior learning and iterative inpainting schemes in SHGM will be introduced in the following two subsections.

### B. SHGM: Prior Learning in Structural-Hankel Domain

Rather than the most common methods that conducting the training process in image domain [30], [57], in this work the image data is constructed into structural-Hankel matrix based tensors for prior learning. The whole training procedure is mainly divided into three steps, as shown in Fig. 4.

***Step 1: Extracting Internal-Middle Patches from Images.*** Unlike traditional generative models, the proposed method only needs a few images for training. To accommodate local image statistics, these original images are randomly cropped into many internal-middle patches **x** with size of $64 \times 64 \times 3$. Note that at each iteration of the training process, an internal-middle patch needs to be extracted from each of these images.

***Step 2: Constructing Hankel Matrices from Internal-Middle Patches.*** We construct the large-scale structural-Hankel matrix on each internal-middle patch. First, the structural-Hankel matrix is constructed on each internal-middle patch with a size of $64 \times 64 \times 3$ by sliding a window with a size of $8 \times 8 \times 3$. Subsequently, we attain a set of structural-Hankel matrix $X = H(\mathbf{x})$ with size of $3249 \times 192$.

***Step 3: Generative Modeling on Folded Tensors.*** Due to the large size of structural-Hankel matrix, the structural-Hankel matrix is folded into higher dimensional tensor (*i.e.*, $\mathcal{X} = T(X)$) for the purpose of efficient score-based modeling. Hence, we obtain a series of folded tensors with size of $192 \times 192 \times 16$. Afterwards, the folded tensors are used as the input of the score-based network for training. Because the structural-Hankel matrix explicitly represents the statistical distribution within images, the redundancy of the same pixel among different positions can be fully exploited to capture the internal statistics distribution from the training samples.

In detail, the process of prior learning in the score-based model is manipulated by using an SDE to transform a complex data distribution into a known prior distribution. Empirically, Variance Exploding (VE) SDE [30] typically leads to higher sample qualities. It can be chosen $f = 0$ and $g = \sqrt{d[\sigma^2(t)]/dt}$ for Eq. (6). The forward VE-SDE can be formulated as:

$$d\mathcal{X} = \sqrt{\frac{d[\sigma^2(t)]}{dt}} dw \qquad (8)$$

where $\sigma(t)$ is Gaussian noise function with a variable in continuous time $t \in [0,1]$, which can be redescribed as a positive noise scale $\{\sigma_i\}_{i=1}^N$. The forward VE-SDE can be described into a Markov chain $\mathcal{X}^{i+1} = \mathcal{X}^i + \sqrt{\sigma_{i+1}^2 - \sigma_i^2}z$, $i = 0, \cdots, N-1$.

Since $\nabla_\mathcal{X} \log p_t(\mathcal{X}(t))$ is intractable for all $t$, it can be estimated by training a score-based model on samples with denoising score matching. To this end, the time-dependent score-based model $s_\theta(\mathcal{X}(t),t)$ can be trained through optimizing the parameters $\theta$:

$$\min_\theta \mathbb{E}_{t \sim U(0,1)} \left\{ \gamma(t) \mathbb{E}_{\mathcal{X}(0)} \mathbb{E}_{\mathcal{X}(t)|\mathcal{X}(0)} \left[ \left\| s_\theta(\mathcal{X}(t),t) - \nabla_\mathcal{X} \log p_{0t}(\mathcal{X}(t) | \mathcal{X}(0)) \right\|_2^2 \right] \right\} \qquad (9)$$

where $\gamma : [0,T] \to \mathbb{R}_{>0}$ is a positive weighting function and $t$ is uniformly sampled over $[0,T]$. $p_{0t}(\mathcal{X}(t) | \mathcal{X}(0))$ is the Gaussian perturbation kernel centered at $\mathcal{X}(0)$. Once the network satisfies $s_\theta(\mathcal{X}(t),t) \simeq \nabla_\mathcal{X} \log p_t(\mathcal{X}(t))$, it means that $\nabla_\mathcal{X} \log p_t(\mathcal{X}(t))$ is known for all $t$ by $s_\theta(\mathcal{X}(t),t)$. Thus, it allows us to derive the reverse-time SDE and simulate it in sample, which can be given by:

$$d\mathcal{X} = -d\sigma^2(t) \cdot \nabla_\mathcal{X} \log p_t(\mathcal{X}) + \sqrt{d[\sigma^2(t)]/dt} d\bar{w} \qquad (10)$$

Comprehensively, the bidirectional process of SDE is described in Fig. 5. Firstly, the forward VE-SDE diffuses the image data distribution $\mathcal{X}(0) \sim p_0$ into a prior distribution $\mathcal{X}(T) \sim p_T$. Secondly, to find the score of the distribution at each intermediate time step, one can estimate a score function with a time-conditional neural network $s_\theta(\mathcal{X}(t),t) \simeq \nabla_\mathcal{X} \log p_t(\mathcal{X}(t))$. Finally, by reversing the forward process and starting from samples of $\mathcal{X}(T) \sim p_T$, samples $\mathcal{X}(0) \sim p_0$ can be obtained.

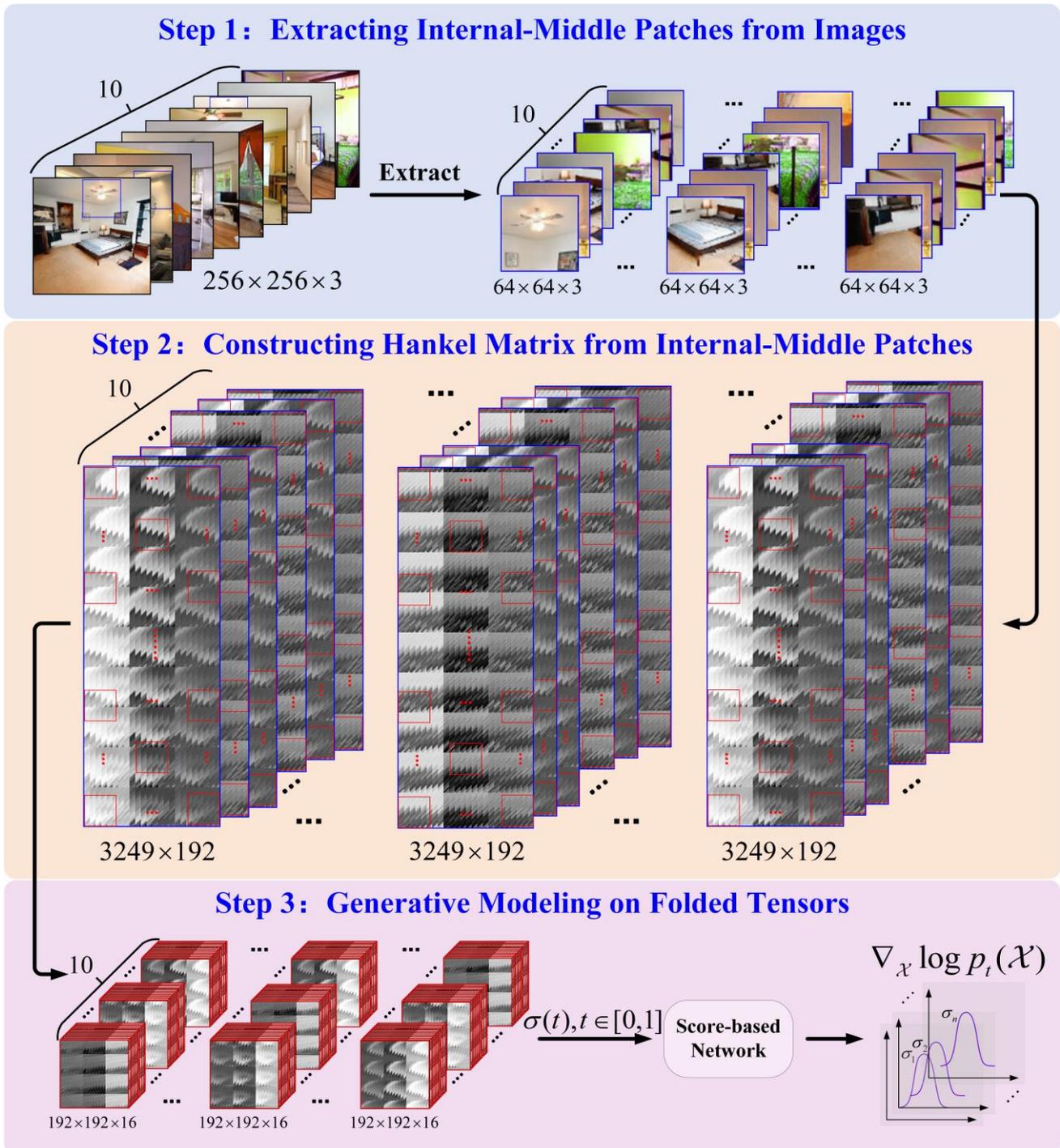

**Fig. 4.** The pipeline of the training process in SHGM. Before the prior learning of the score-based network, internal-middle patches are extracted from the original images and then transformed into structural-Hankel matrices as the training samples. After the training procedure is finished, the network $s_\theta(\mathcal{X}(t),t)$ learns the data score $\nabla_\mathcal{X} \log p_t(\mathcal{X})$.

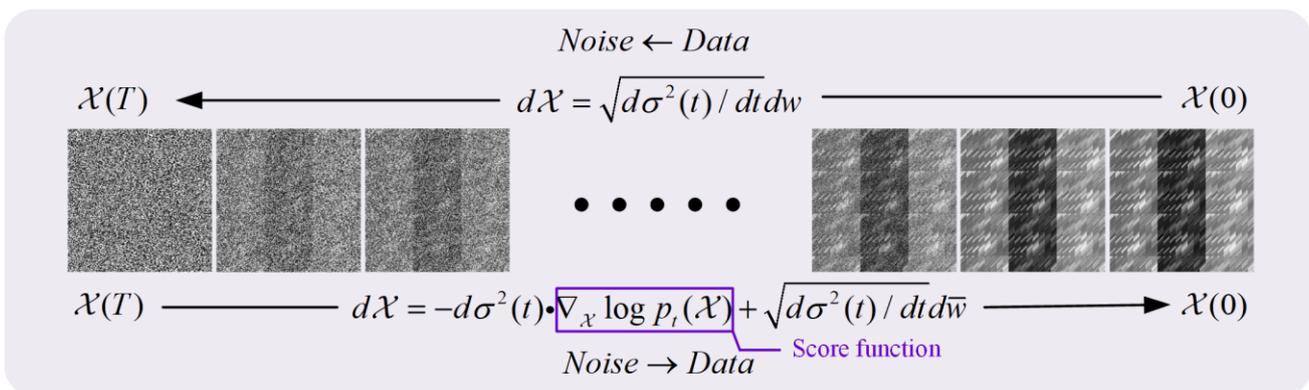

**Fig. 5.** Forward and reverse-time process of VE-SDE on structural-Hankel patches.

***Step 1: Extracting Internal-Middle Patches.*** To effectively handle the boundary of the image, the observation is padded to be the size of $320 \times 320 \times 3$. Subsequently, there are 25 patches of size $64 \times 64 \times 3$ (*i.e.*, $\{\mathbf{x}_n, n=1,\cdots,25\}$), which are extracted from the padded image in patch-by-patch manner.

***Step 2: Constructing Hankel Matrices on Patches.*** The patches obtained in the above ***Step 1*** are sequentially constructed into structural-Hankel matrices, and the implementation is similar to ***Step 2*** in the training process. Whereafter, thees large-scale structural-Hankel matrices with size of $3249 \times 192$ are transformed to be tensors with suitable size for restoration.

***Step 3: Conditional Generation.*** At the conditional generation process, the PC sampler, ADMM, and DC steps are alternatively performed. Afterwards, the restored image is obtained by concatenating the updated patches.

The forward VE-SDE process allows us to produce data samples not only from $p_0$, but also from $p_0[(\mathcal{X}(0), \mathcal{X}_{LR})|Y]$ if $p_t[Y|(\mathcal{X}, \mathcal{X}_{LR})]$ is known. Thus, we can sample from $p_t[(\mathcal{X}, \mathcal{X}_{LR})|Y]$ by starting from $p_T[(\mathcal{X}(T), \mathcal{X}_{LR})|Y]$. In other words, the proposed SHGM views the observation as a conditional generation in a low-rank environment and incorporates it into the iterative inpainting procedure, i.e.,

$$\nabla_\mathcal{X} \log p_t[(\mathcal{X}, \mathcal{X}_{LR})|Y] \simeq \nabla_\mathcal{X} \big[\log p_t(\mathcal{X}|\mathcal{X}_{LR}) + \log p_t(\mathcal{X}_{LR}) \\ + \log p_t[Y|(\mathcal{X}, \mathcal{X}_{LR})]\big] \quad (11)$$

where $Y$ is the measurement corresponding to $\mathcal{X}$. Then, plugging Eq. (11) into the counterpart of Eq. (10), the conditional sampling by solving a conditional reverse-time VE-SDE, whose discretization is expressed as:

$$\mathcal{X}_n^i = \mathcal{X}_n^{i+1} + (\sigma_{i+1}^2 - \sigma_i^2)\nabla_\mathcal{X}\big[\log p_t(\mathcal{X}_n^{i+1}|\mathcal{X}_{LR}^{i+1}) + \log p_t(\mathcal{X}_{LR}^{i+1}) \\ + \log p_t[Y_n|(\mathcal{X}_n^{i+1}, \mathcal{X}_{LR}^{i+1})]\big] + \sqrt{\sigma_{i+1}^2 - \sigma_i^2}z \quad (12)$$

Eq. (12) can be decoupled into three sub-problems: The measurement distribution term with low-rank conditional $\log p_t(\mathcal{X}|\mathcal{X}_{LR})$, the prior information related to the structural-Hankel matrix $\log p_t(\mathcal{X}_{LR})$, and the DC $\log p_t[Y|(\mathcal{X}, \mathcal{X}_{LR})]$. Consequently, the conditional generation can be performed by alternatively iterating the PC sampler, ADMM, and DC. The details are demonstrated in following steps:

***Step 3-1: PC Sampler.*** According to Section III. B, $s_\theta(\mathcal{X}(t),t)$ trained with denoising score matching almost satisfies $s_\theta(\mathcal{X}(t),t) \simeq \nabla_\mathcal{X} \log p_t(\mathcal{X}|\mathcal{X}_{LR})$. The PC sampler is introduced to correct errors in the evolution of the discretized reverse-time SDE.

Specifically, the predictor refers to a numerical solver for the reverse-time SDE, which is used for giving an estimate of the sample at the next time step:

$$\mathcal{X}_n^i = \mathcal{X}_n^{i+1} + (\sigma_{i+1}^2 - \sigma_i^2)s_\theta(\mathcal{X}_n^{i+1},t) + \sqrt{\sigma_{i+1}^2 - \sigma_i^2}z \quad (13)$$

where $i = N-1,\cdots,1,0$ is the number of discretization steps for the reverse-time SDE, $\sigma_i$ is the noise schedule at the $i^{th}$ iteration, and $z \sim \mathbb{N}(0,1)$ is the standard normal.

Meanwhile, the corrector is defined as the iteration procedure of Langevin dynamics, which is used for correcting the marginal distribution of the estimated sample:

$$\mathcal{X}_n^{i,j} = \mathcal{X}_n^{i,j-1} + \varepsilon_i s_\theta(\mathcal{X}_n^{i,j-1},t) + \sqrt{2\varepsilon_i}z \quad (14)$$

where $j = 1,2,\cdots,M$ is the number of corrector steps, $\varepsilon_i > 0$ is the noise step size at the $i^{th}$ iteration. During the above iterative procedure, when $N \to \infty$, $M \to \infty$ and $\varepsilon_i \to 0$, $\mathcal{X}_n^{i,j}$ is a sample from $p_t(\mathcal{X}_{LR})$ under some regularity conditions.

***Step 3-2: ADMM.*** After PC sampling, the updated $\mathcal{X}_n^i$ is again unfolded to be a structural-Hankel matrix of size $3249 \times 192$. According to the low-rank property in structural-Hankel matrix, the formulation of low-rank constraint term in Eq. (12) is described as:

$$\min_{X_{LR}} rank(X_{LR}) \quad s.t. \quad X_{LR} = T^{-1}(\mathcal{X}_n) \quad (15)$$

Subsequently, ADMM [53] is elaborated to handle Eq. (15), which is easy to obtain:

$$U = \mu(T^{-1}(\mathcal{X}_n) + \Lambda)V(I + \mu V^H V)^{-1} \quad (16)$$

$$V = \mu(T^{-1}(\mathcal{X}_n) + \Lambda)^H U(I + \mu U^H U)^{-1} \quad (17)$$

$$\Lambda = T^{-1}(\mathcal{X}_n) - UV^H + \Lambda \quad (18)$$

$$X_{LR} = U_n V^H - \Lambda \quad (19)$$

where $\mu$ is an averaged compression ratio at numerical ranks, H is the matrix transpose operator, $T^{-1}$ is the unfold operator, $X_{LR}$ is updated from ADMM. To reduce the computational complexity, we initialize $U, V, \Lambda$ via introducing another SVD-free algorithm named LMaFit [46].

***Step 3-3 DC.*** After each PC sampler and ADMM iteration step, the DC step is performed on the intermediate data. In particular, after updating the virtual variable $X_{LR}$ via ADMM, $X_{LR}$ is turned back to the image patch form via the reverse operator $H^\dagger$. Plugging it into the DC formulation, it yields:

$$\mathbf{x}_n^i = \arg\min_{\mathbf{x}_n}\big[\|D\mathbf{x}_n - y_n\|^2 + \lambda\|\mathbf{x}_n - H^\dagger(X_{LR})\|^2\big] \quad (20)$$

The least-square minimization in Eq. (20) can be solved as follows:

$$\mathbf{x}_n^i = \frac{D^T y_n + \lambda H^\dagger(X_{LR})}{1 + \lambda} \quad (21)$$

Comprehensively, the prior learning and iterative inpainting pseudo-code are formally described in **Algorithm 1**. The whole iterative inpainting procedure consists of a two-level loop: The outer loop and the inner loop perform predictor and corrector, respectively. At the same time, ADMM and DC steps are incorporated into both the outer loop and the inner loops, respectively. Fig. 6 visualizes the iterative inpainting process of SHGM for recovering a $256 \times 256 \times 3$ image. More precisely, step 1 is the process of extracting patches from the observation after padding. Step 2 constructs these patches into structural-Hankel matrices of size $3249 \times 192$ and then folds them into tensors with a size of $192 \times 192 \times 16$. Subsequently, step 3 starts conditional generation. Firstly, the tensors obtained in step 2 are updated by PC sampler. Secondly, the updated tensors of size $192 \times 192 \times 16$ are again unfolded into structural-Hankel matrices of size $3249 \times 192$ and then these matrices are then updated alternatively through ADMM. Subsequently, intermediate patches are restored by the reverse operator $H^\dagger$ from the updated structural-Hankel matrices. Finally,

DC is plugged into these intermediate patches to obtain $\mathbf{x}_n^i$ as the input for the next iteration. When the iteration is over, the final restored image $x_{rec}$ is acquired by concatenating all the patches { $\mathbf{x}_n^0$, $n = 1,\cdots, 25$ }.

---

**Algorithm 1 SHGM**

**Generative modeling for prior learning**

**Dataset:** Several image data $x$

1: Extracting internal-middle patches $\mathbf{x}$
2: Constructing Hankel matrices $X = H(\mathbf{x})$
3: Training $s_\theta(\mathcal{X},t) \simeq \nabla_\mathcal{X} \log p_t(\mathcal{X})$ on folded tensors $\mathcal{X} = T(X)$

**Output:** Trained $s_\theta(\mathcal{X},t)$

**Conditional generation for iterative inpainting**

**Initialize:** $\sigma_i, \varepsilon_i, T, N, M, U, V, \Lambda$

1: $\mathcal{X}_n = T[H(\mathbf{x}_n)]$; $\mathcal{X}_n^\mathbb{N} \sim \mathbb{N}(0, \sigma_T \mathbf{I})$
2: **For** $i = N-1$ **to** 0 **do (Outer loop)**
3:    Update $\mathcal{X}_n^i$ via Eq. (13) **(Predictor)**
4:    Update $U^i, V^i, \Lambda^i, X_{LR}^i$ via Eqs. (16-19) **(ADMM)**
5:    Update $\mathbf{x}_n^i$ via Eq. (21) **(DC)**
6:    **For** $j = 1$ **to** $M$ **do (Inner loop)**
7:      Update $\mathcal{X}_n^{i,j}$ via Eq. (14) **(Corrector)**
8:      Update $U^{i,j}, V^{i,j}, \Lambda^{i,j}, X_{LR}^{i,j}$ via Eqs. (16-19) **(ADMM)**
9:      Update $\mathbf{x}_n^{i,j}$ via Eq. (21) **(DC)**
10:   **End for**
11: **End for**
12: $x_{rec} = Concatenate(\mathbf{x}_n^0)$ $n = 1,2,\cdots, 25$

**Return** $x_{rec}$

---

## IV. EXPERIMENTS

In this section, a set of experiments on color image restoration are presented to demonstrate the effectiveness, robustness, and flexibility of SHGM. Particularly, the experimental implementations and datasets for evaluation are detailed.

### A. Experiment Setup

**1) Datasets:** All the image inpainting tests are conducted on LSUN dataset and seven standard natural images [35]. Specifically, the seven standard natural images consist of Baboon, Barbara, Boat, Cameraman, House, Lena, and Peppers. These images are rich in color and texture, which is ideal for viewing restoration results. Meanwhile, the LSUN dataset is a large color image dataset that encompasses around one million labeled images for each of 10 scene categories and 20 object categories. Among them, we choose the indoor scene LSUN-bedroom dataset that has enough samples (more than 3 million) and various colors to verify the effectiveness of SHGM. It is worth noting that we use a model trained on only 10 bedroom images for testing on 7 standard natural images and 100 bedroom images, respectively. Besides, the classic large-scale datasets ImageNet and Berkeley Segmentation Dataset (BSD), consisting of various images ranging from natural images to specific objects such as plants, people, food *etc.*, are selected to demonstrate the remarkable performance and diversity of SHGM.

**2) Parameter Setting:** The pixel values of all images are normalized to the range $[0,1]$. The data distribution is perturbed with Gaussian noise with noise values ranging from $0.01 \sim 378$. Using a fixed learning rate of $0.0002$, the optimization adopts Adaptive Moment Estimation. In the iterative inpainting phase, the number of outer iterations is set to $N = 1000$ and the number of inner iterations is $M = 1$. Besides, the signal-to-noise ratio is $SNR = 0.21$. Each time the prediction process of the outer loop is performed, the correction process of the inner loop is iterated once by annealing Langevin. We adopt the architecture of NCSN++. A major difference from the work in [30] is that its network input is three-channel image, while SHGM leverages sixteen-channel tensors.

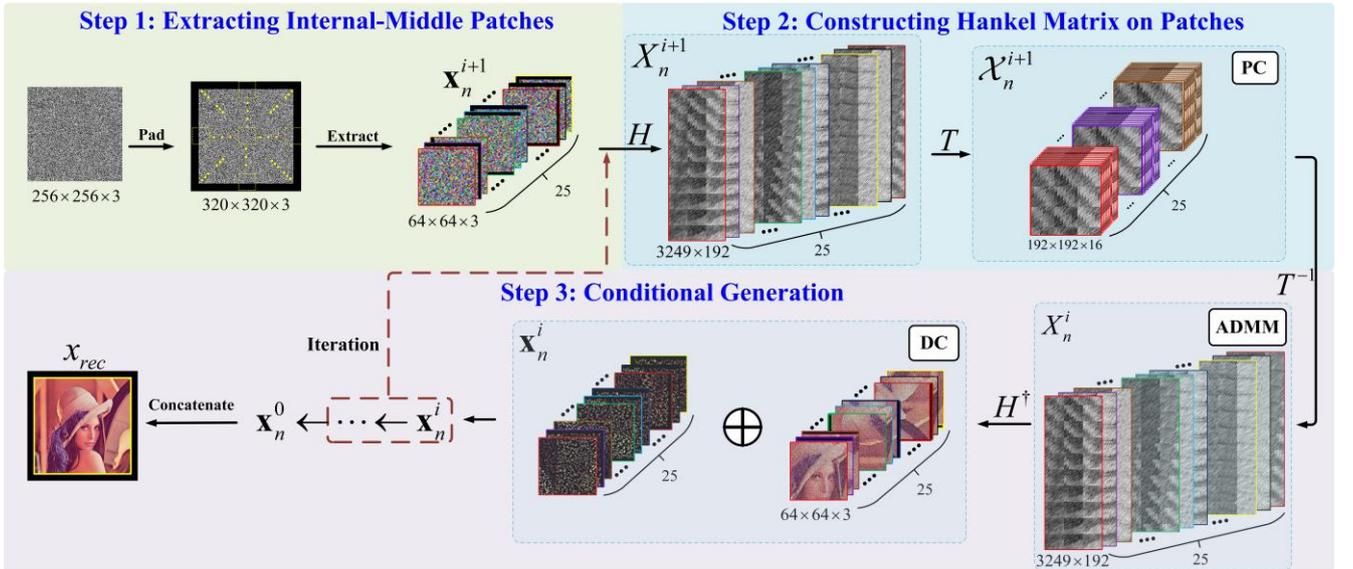

**Fig. 6.** The pipeline of iteration inpainting procedure in SHGM, which mainly consists of three steps. Firstly, patches are extracted patch-by-patch from the observation after padding. Then, constructing structural-Hankel matrices on patches and folding them with the size of $192 \times 192 \times 16$. Finally, the PC sampler, ADMM, and DC steps are performed alternatively.

**3) Evaluation Metrics:** In all reported experiments, two traditional metrics -- peak signal to noise ratio (PSNR) and structural similarity (SSIM) are recorded to evaluate the quality of images restored by different approaches. The higher PSNR and SSIM values, the better visual quality with more details. Denoting $x$ and $\hat{x}$ to be the recon-

structed image and the ground truth, the PSNR is defined as:
$$PSNR(x,\hat{x}) = 20\log_{10}[Max(\hat{x})/\|x-\hat{x}\|_2] \quad (22)$$

The SSIM is defined as:
$$SSIM(x,\hat{x}) = \frac{(2\mu_x\mu_{\hat{x}}+c_1)(2\sigma_{x\hat{x}}+c_2)}{(\mu_x^2+\mu_{\hat{x}}^2+c_1)(\sigma_x^2+\sigma_{\hat{x}}^2+c_2)} \quad (23)$$

The training and experiments are performed with a customized version of Pytorch on an Intel i7-6900K CPU and a GeForce Titan XP GPU. For the convenience of reproducible research, the source code of SHGM can be downloaded from the website: https://github.com/yqx7150/SHGM.

### B. Experimental Comparison

In this subsection, to reveal the remarkable performance and diversity of SHGM, we test it on LSUN-bedroom dataset and seven standard natural images, respectively.

**1) Test on LSUN-bedroom Dataset:** In this experiment, three masks are chosen as the degradation operators: Block, text, and random. To verify that SHGM can learn generalizable and semantically meaningful image representations, we compare it with Kernel (steering) [54], K-SVD [55], RFR [56], ALOHA [35], and NCSN++ [30] with respect to image recovery results.

Fig. 7 provides the comparative results of inpainting experiments on randomly removing 90% pixels in image. It is intuitive from the global comparison that other methods still retain varying degrees of artifacts and noises. Although ALOHA performs well at random mask in terms of PSNR value, its visual effect is far inferior to that of SHGM. As can be seen in the magnified area, the traces are still remaining on the white sheet. ALOHA and SHGM work much better than NCSN++. The results of ALOHA are too aggregated, resulting in an overly smooth sheet texture. Clearly, the SHGM results are better at restoring texture. Fig. 8 demonstrates a set of comparison results restored from the image under the block covering 50%. Obviously, the results of RFR still remain traces of the block. Both NCSN++ and SHGM perform more satisfactorily. Observing the magnified area, SHGM generates more semantically plausible and visual-pleasing results.

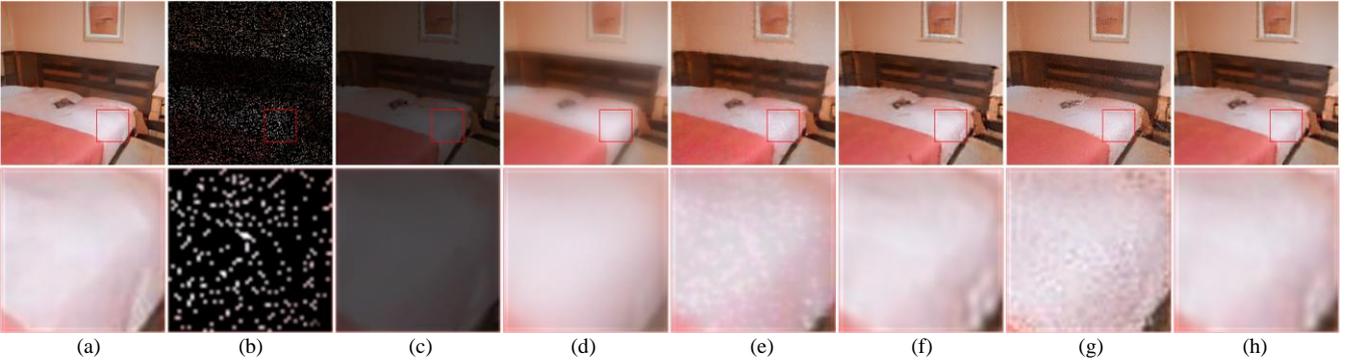

Fig. 7. Subjective comparison on LSUN-bedroom dataset. The results are restored from the observation under the 90% missing samples. (a) Ground truth, (b) Observation, (c) Kernel, (d) K-SVD, (e) RFR, (f) ALOHA, (g) NCSN++, (h) SHGM.

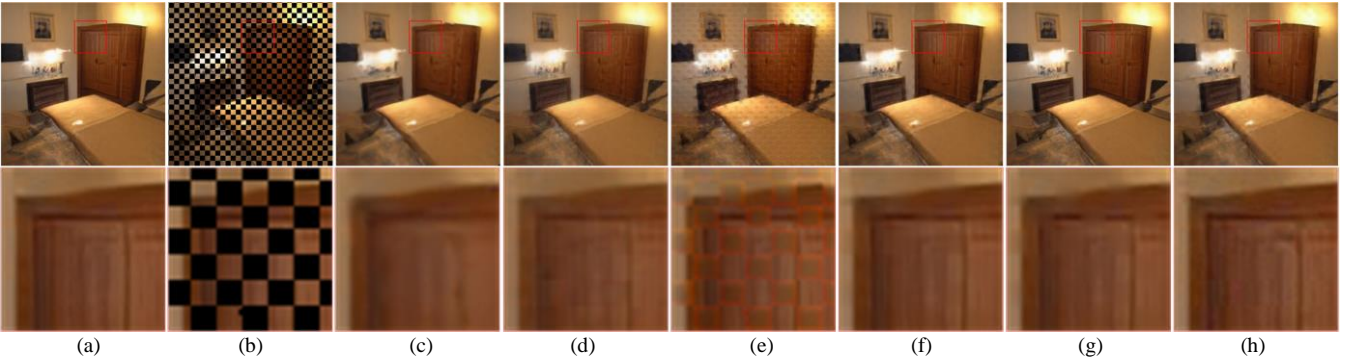

Fig. 8. Subjective comparison on LSUN-bedroom dataset. The results are restored from the observation, which covers block with 50%. (a) Ground truth, (b) Observation, (c) Kernel, (d) K-SVD, (e) RFR, (f) ALOHA, (g) NCSN++, (h) SHGM.

Table I reports the average PSNR and SSIM values over one hundred images for each algorithm from block, text, and random masks. Apparently, SHGM outperforms the other comparative algorithms in most cases. To be specific, the PSNR value of SHGM is 4.59 dB higher than ALOHA in the case of Text2 and 3.23 dB higher than NCSN++ in terms of missing 80% pixels. Similarly, the SSIM value is another metric to assess the performance of natural image inpainting. SHGM retains a higher SSIM under following two text and three random masks. Totally, the qualitative and quantitative results exhibit that the overall effect of SHGM is superior to that of all the other compared methods. The images generated by our proposed model are closer to the ground truth than those generated by other methods. Emphatically,
NCSN++ utilizes tens of thousands of images with plentiful diversities in the training stage. However, we merely select ten images of bedroom from LUSN dataset, and the outcomes of SHGM outperform NCSN++.

**2) Test on Seven Standard Natural Images:** To evaluate the generalization of prior knowledge learned by SHGM from LUSN-bedroom dataset, we conduct inpainting experiments on Baboon, Barbara, Boat, Cameraman, House, Lena, and Peppers with 80% (or 50%) of pixels randomly removed. Kernel (steering) [54], K-SVD [55], and ALOHA [35] are chosen as the comparative algorithms.

Here, Figs. 9 and 10 exhibit restored images of Baboon and Lena, respectively. Compared with the restoration effects of other methods, the restored image texture structure of SHGM is clearer with seldom artifacts and noises. It is obvious that SHGM is able to generate realistic images that closely resemble the ground truth. Apart from that, SHGM maintains fine features in the edge details. It is most comparable in terms of qualitative visual quality. The comparison of image quality metrics for all inpainting methods indicates the superior performance of SHGM.

TABLE I
RESTORATION PSNR AND SSIM VALUES BY VARIOUS IMAGE INPAINTING ALGORITHMS FROM BLOCK, TEXT, AND RANDOM MASKS.

| Algorithm | | Kernel (steering) [54] | K-SVD [55] | RFR [56] | ALOHA [35] | NCSN++ [30] | SHGM |
|---|---|---|---|---|---|---|---|
| Block | | 18.11/0.7191 | 27.15/0.8600 | 24.64/0.6981 | 27.82/**0.8792** | 27.27/0.8556 | **28.08**/0.8759 |
| Text | Text1 | 26.78/0.8435 | 33.75/0.9641 | 29.46/0.8793 | 34.08/0.9659 | 33.60/0.9569 | **34.33/0.9660** |
| | Text2 | 15.64/0.6567 | 35.55/0.9732 | 33.09/0.9600 | 31.87/0.9710 | **36.58**/0.9807 | 36.46/**0.9816** |
| Random | 90% | 12.01/0.5464 | 22.51/0.6753 | 24.22/0.7160 | 24.65/0.7639 | 20.65/0.5185 | **25.57/0.7881** |
| | 80% | 24.69/0.8154 | 26.27/0.8093 | 25.70/0.7504 | 27.58/0.8540 | 25.19/0.7550 | **28.42/0.8702** |
| | 70% | 26.23/0.8305 | 27.72/0.8607 | 25.86/0.7507 | 29.56/0.8994 | 27.52/0.8332 | **29.91/0.9025** |

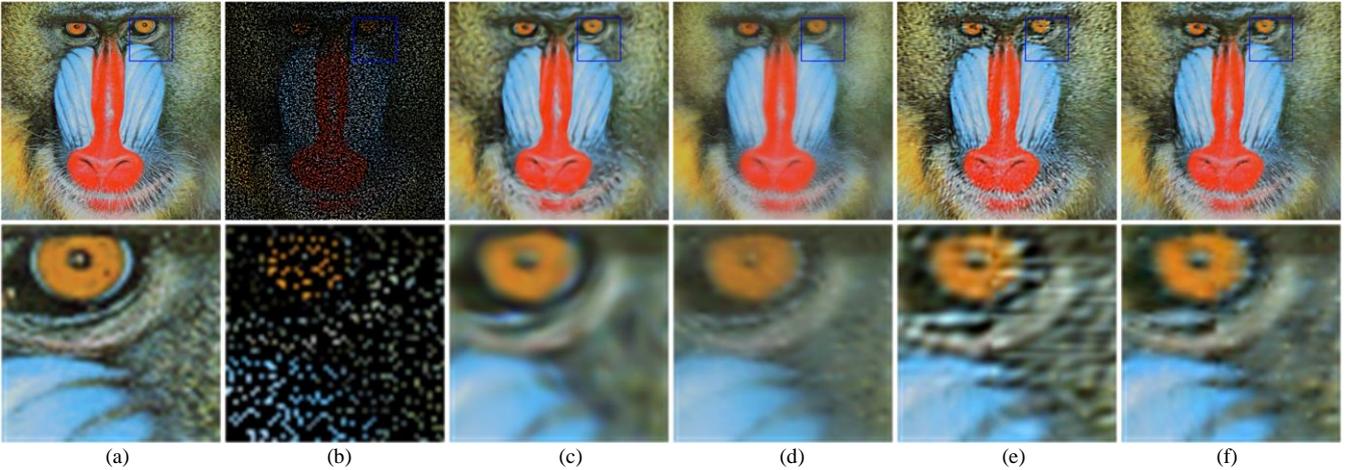

**Fig. 9.** Subjective comparison for restoring image Baboon. The results are restored from the observation under 80% missing samples. (a) Ground truth, (b) Observation, (c) Kernel, (d) K-SVD, (e) ALOHA, (f) SHGM.

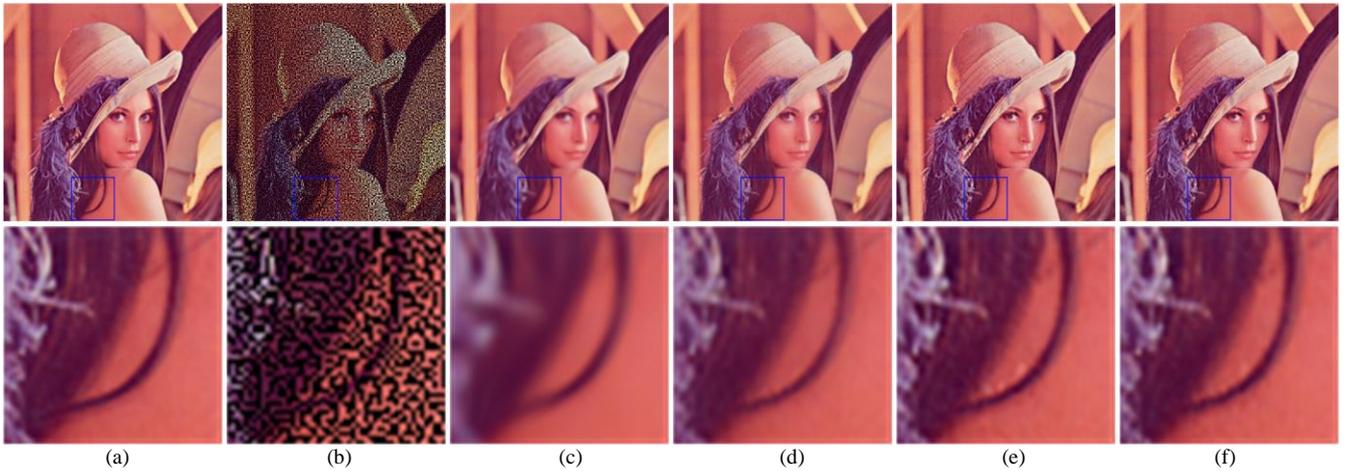

**Fig. 10.** Subjective comparison for recovering image Lena. The results are restored from the observation under 50% missing samples. (a) Ground truth, (b) Observation, (c) Kernel, (d) K-SVD, (e) ALOHA, (f) SHGM.

As shown in Table II, the PSNR value of SHGM is 1.63 dB higher than ALOHA for filling in image Baboon with 80% missing pixels. Meanwhile, the SSIM value of SHGM attains 0.6495, which is also the optimal among all comparison methods. For the task of recovering 50% pixels for Peppers, the PSNR and SSIM values obtained by SHGM are the highest among the four methods. Additionally, the PSNR value of SHGM is higher than Kernel, K-SVD, and ALOHA with 5.37 dB, 3.28 dB, and 0.75 dB, respectively. Therefore, SHGM is a competitive method owning a better restoration effect. Especially, in other pictures with less texture and less color difference, SHGM can achieve remarkable restoration results. Unfortunately, since Barbara owns the complex textures and sufficient color, the visual effect of SHGM is not obvious with ALOHA.

Comprehensively, it should be emphasized that the seven standard natural images are tested on SHGM, which is trained from ten images of bedroom. Although the seven standard natural images are severely different from bedroom images in visual scene and style, the restoration performance remains stable.

### C. Ablation Study

In the present SHGM, a small amount of training data and training in structural-Hankel domain are two key factors that lead to flexible and accurate restoration. In this subsection, the influence of these factors will be investigated.

TABLE II
RESTORATION PSNR AND SSIM VALUES BY VARIOUS IMAGE INPAINTING ALGORITHMS FROM 80% AND 50% MISSING SAMPLES.

| Test Image / Algorithm | | Kernel (steering) [54] | K-SVD [55] | ALOHA [35] | SHGM |
|---|---|---|---|---|---|
| Baboon | 80% | 21.90/0.5460 | 22.52/0.5694 | 20.91/0.5993 | **22.54/0.6495** |
|  | 50% | 22.75/0.5817 | 25.71/0.8163 | 24.75/0.8169 | **26.14/0.8522** |
| Barbara | 80% | 26.73/0.8151 | 26.08/0.7880 | 27.51/0.8501 | **28.28/0.8652** |
|  | 50% | 27.98/0.8429 | 30.19/0.9076 | **33.59/0.9566** | 32.57/0.9453 |
| Boat | 80% | 24.91/0.7475 | 25.22/0.7591 | 25.99/0.8144 | **26.26/0.8204** |
|  | 50% | 26.18/0.7867 | 30.06/0.9188 | 31.42/0.9397 | **31.95/0.9453** |
| Cameraman | 80% | 23.93/0.7965 | 23.92/0.7893 | 24.68/0.8263 | **24.95/0.8298** |
|  | 50% | 25.45/0.8246 | 28.55/0.9122 | 29.53/0.9255 | **29.98/0.9306** |
| House | 80% | 28.48/0.8197 | 27.67/0.8173 | 31.91/0.8723 | **32.22/0.8786** |
|  | 50% | 30.40/0.8429 | 32.85/0.9270 | 37.60/0.9499 | **37.91/0.9538** |
| Lena | 80% | 28.57/0.8605 | 26.98/0.8250 | 28.25/0.8713 | **29.29/0.8873** |
|  | 50% | 29.82/0.8807 | 31.75/0.9342 | 33.91/0.9514 | **34.49/0.9577** |
| Peppers | 80% | 27.75/0.8819 | 26.18/0.8419 | 28.17/0.8804 | **28.83/0.8917** |
|  | 50% | 29.17/0.9018 | 31.26/0.9380 | 33.79/0.9496 | **34.54/0.9578** |

*1) Different Number of Raw Training Data of SHGM:* To compare the effect of different numbers of raw training data on the restored results, we select 1, 3, 5, and 10 images for comparative experimental.

Two quantitative metrics of the restoration are listed in Table III. Under the premise with randomly removing 80% pixels from Barbara, the highest PSNR value of 28.28 dB is obtained with 10 images as training. The SSIM value is 0.8652, which is 0.0055 away from the lowest value. In fact, Table III shows that there is little difference in the test results for varying numbers of training images. The integral effect of the two quantitative metrics tested using 10 training images is slightly better. Therefore, it can be seen from this phenomenon that SHGM can learn effective internal statistical information within images by training the model with only several or even one image represented in the Hankel domain. In addition, the input number of training images has little influence on time consumption. Hence, all the experiments in our work select 10 images as the network input under a comprehensive trade-off.

TABLE III
RESTORATION PSNR AND SSIM VALUES ON 80% MISSING SAMPLES BY USING DIFFERENT NUMBER OF RAW TRAINING DATA.

| Number | 1 | 3 | 5 | 10 |
|---|---|---|---|---|
| Barbara | 28.12/0.860 | 28.10/0.860 | 28.25/0.865 | **28.28/0.865** |
| Peppers | 28.82/0.889 | **28.87**/0.891 | 28.83/0.891 | 28.83/**0.892** |

*2) Different Training Schemes of SHGM:* To assess the superiority of SHGM, we conduct the experiment to compare SHGM with another training scheme which is trained in image domain, i.e., the training scheme in image domain does not require the Hankel transformation and trains internal-middle patches to learn the prior distribution.

Quantitative metrics are listed in Table IV. Under the premise with randomly removing 80% pixels from Lena, SHGM achieves the PSNR value that is 0.15 dB higher than the training scheme in image domain. The SSIM value of SHGM reaches 0.8873. It can be inferred that the advantage of training in structural-Hankel domain is making full use of the low-rank and redundant properties. Thus, it can effectively remedy for the shortcomings of insufficient training in image domain.

TABLE IV
RESTORATION PSNR AND SSIM VALUES ON 80% MISSING SAMPLES BY USING DIFFERENT TRAINING SCHEMES.

| Training domain | Image domain | Hankel domain |
|---|---|---|
| Bedroom | 28.13/0.8652 | **28.42/0.8702** |
| Lena | 29.14/0.8775 | **29.29/0.8873** |

*3) Different Training Datasets of SHGM:* Here, we train two SHGM models from 10 images that selected from BSD and ImageNet datasets, respectively. Then, 100 images from Bedroom dataset are used to evaluate these two models. As reported in Table V, the restoration PSNR and SSIM values under the model learned from ImageNet are slightly better than that from BSD dataset. Accordingly, SHGM can indeed achieve comparable results under different training datasets and merely using several raw training data.

TABLE V
RESTORATION PSNR AND SSIM VALUES BY BSD AND IMAGENET DATASETS FROM BLOCK MASKS.

| Training dataset | BSD | ImageNet |
|---|---|---|
| 100 images | 27.86/0.8689 | **28.02/0.8747** |

## IV. CONCLUSION

This work proposed a structural-Hankel matrix-assisted score-based generative model SHGM for color image inpainting. The proposed SHGM collected several images to construct sufficient samples for prior learning and the structural-Hankel domain was explored to assist score-based model learn the distribution inherent in the image. As the prior knowledge was determined, it can be incorporated into the conditional generation for various inpainting tasks. Experimental results verified that SHGM can produce outstanding performance under different masks with different datasets, keeping higher PSNR and SSIM values than state-of-the-arts. More application scenarios will be exploited in the future study to validate the robustness and flexibility of the presented methodology.

REFERENCE

[1] D. Pathak, P. Krahenbuhl, J. Donahue, T. Darrell, and A. A. Efros, "Context encoders: Feature learning by inpainting," *Proc. IEEE Conf. Comput. Vis. Pattern Recognit.*, pp. 2536-2544, Mar. 2016.
[2] G. Liu, F. A. Reda, K. J. Shih, T. C. Wang, A. Tao, and B. Catanzaro, "Image inpainting for irregular holes using partial convolutions," *Proc. Eur. Conf. Comput. Vis.*, pp. 85-100, 2018.

[3] Z. Yi, Q. Tang, S. Azizi, D. Jang, and Z. Xu, "Contextual residual aggregation for ultra high-resolution image inpainting," *Proc. IEEE Conf. Comput. Vis. Pattern Recognit.*, pp. 7508-7517, 2020.

[4] W. Du, H. Chen, and H. Yang, "Learning invariant representation for unsupervised image restoration," *Proc. IEEE Conf. Comput. Vis. Pattern Recognit.*, pp. 14483-14492, 2020.

[5] Z. Wan, B. Zhang, D. Chen, P. Zhang, D. Chen, J. Liao, and F. Wen, "Bringing old photos back to life," *Proc. IEEE Conf. Comput. Vis. Pattern Recognit.*, pp. 2747-2757, 2020.

[6] D. Ulyanov, A. Vedaldi, and V. Lempitsky, "Deep image prior," *Proc. IEEE Conf. Comput. Vis. Pattern Recognit.*, vol. 1, pp. 9446-9454, 2018.

[7] X. Ning, W. Li, and W. Liu, "A fast single image haze removal method based on human retina property," *IEICE Trans. Inf. Syst.*, vol. 100, no. 1, pp. 211-214, Jan. 2017.

[8] M. Bertalmio, G. Sapiro, V. Caselles, and C. Ballester, "Image inpainting," *Proc. 27th Annu. Conf. Comput. Graph. Interact. Techn.*, pp. 417-424, Jul. 2000.

[9] C. Ballester, M. Bertalmio, V. Caselles, G. Sapiro, J. Verdera, "Filling-in by joint interpolation of vector fields and gray levels," *IEEE Trans. Image Process.*, vol. 10, no. 8, pp. 1200-1211, Aug. 2001.

[10] A. Levin, A. Zomet, and Y. Weiss, "Learning how to inpaint from global image statistics," *Proc. 9th IEEE Int. Conf. Comput. Vis.*, vol. 1, pp. 305-312, Oct. 2003.

[11] S. Esedoglu, and J. Shen, "Digital inpainting based on the Mumford-Shah-Euler image model," *Eur. J. Appl. Math.*, vol. 13, no. 4, pp. 353-370, Aug. 2002.

[12] M. Bertalmio, L. Vese, G. Sapiro, and S. Osher, "Simultaneous structure and texture image inpainting," *IEEE Trans. Image process.*, vol. 12, no. 8, pp. 882-889, Aug. 2003.

[13] A. Criminisi, P. Perez, and K. Toyama, "Object removal by exemplar-based inpainting," *2003 IEEE Comput. Soc. Conf. Comput. Vis. Pattern Recognit.*, vol. 2, pp. II-II, Jun. 2003.

[14] A. Criminisi, P. Perez, K. Toyama, "Region filling and object removal by exemplar-based image inpainting," *IEEE Trans. Image Process.*, vol. 13, no. 9, pp. 1200-1212, Sep. 2004.

[15] C. Barnes, E. Shechtman, A. Finkelstein, and D. B. Goldman, "Patchmatch: A randomized correspondence algorithm for structural image editing," *ACM Trans. Graph.*, vol. 28, no. 3, pp. 24, Aug. 2009.

[16] J. B. Huang, S. B. Kang, N. Ahuja, and J. Kopf, "Image completion using planar structure guidance," *ACM Trans. Graph.*, vol. 33, no. 4, pp. 1-10, Jul. 2014.

[17] I. Goodfellow, J. Pouget-Abadie, M. Mirza, B. Xu, D. Warde-Farley, S. Ozair, A. Courville, and Y. Bengio, "Generative adversarial nets," *Adv. Neural Inf. Process. Syst.*, pp. 2672-2680, 2014.

[18] M. Mirza, and S. Osindero, "Conditional generative adversarial nets," *arXiv preprint arXiv:* 1411.1784, 2014.

[19] A. Brock, J. Donahue, and K. Simonyan, "Large scale GAN training for high fidelity natural image synthesis," *arXiv preprint arXiv*: 1809.11096, 2018.

[20] T. R. Shaham, T. Dekel, and T. Michaeli, "Singan: learning a generative model from a single natural image," *Proc. IEEE Int. Conf. Comput. Vis.*, pp. 4570-4580, 2019.

[21] T. Hinz, M. Fisher, O. Wang, and S. Wermter, "Improved techniques for training single-image gans," *Proc. IEEE Winter Conf. Appl. Comput. Vis.*, pp. 1300-1309, 2021.

[22] V. Sushko, J. Gall, and A. Khoreva, "One-shot gan: Learning to generate samples from single images and videos," *Proc. IEEE Conf. Comput. Vis. Pattern Recognit.*, pp. 2596-2600, 2021.

[23] Z. Zheng, J. Xie, and P. Li, "Patchwise generative convnet: Training energy-based models from a single natural image for internal learning," *Proc. IEEE Conf. Comput. Vis. Pattern Recognit.*, pp. 2961-2970, 2021.

[24] Z. Gao, G. Zhai, H. Deng, X. Yang, "Extended geometric models for stereoscopic 3D with vertical screen disparity," *Displays*, vol. 65, Dec. 2020.

[25] J. Zhang, Z. Xie, J. Sun, X. Zou, "A cascaded R-CNN with multiscale attention and imbalanced samples for traffic sign detection," *IEEE Access*, vol. 8, pp. 29742-29754, 2020.

[26] J. Zhang, W. Wang, C. Lu, J. Wang, A. K. Sangaiah, "Lightweight deep network for traffic sign classification," *Ann. Telecommun.*, vol. 75, no. 7, pp. 369-379, Jul. 2019.

[27] C. Yan, G. Pang, X. Bai, *et al.* "Beyond triplet loss: Person re-identification with fine-grained difference-aware pairwise loss," *IEEE Trans. Multimedia*, vol. 24, pp. 1665-1677, Mar. 2021.

[28] Y. Song, and S. Ermon, "Generative modeling by estimating gradients of the data distribution," *Adv. Neural Inf. Process. Syst.*, pp. 11895-11907, 2019.

[29] Y. Song, and S. Ermon, "Improved techniques for training score-based generative models," *Adv. Neural Inf. Process.Syst.*, vol. 33, pp. 12438-12448, 2020.

[30] Y. Song, J. Sohl-Dickstein, D. P. Kingma, *et al.* "Score-based generative modeling through stochastic differential equations," *Int. Conf. Learn. Representations*, 2021.

[31] J. Yu, Z. Lin, J. Yang, X. Shen, X. Lu, T. S. Huang, "Generative image inpainting with contextual attention," *Proc. IEEE Conf. Comput. Vis. Pattern Recognit.*, pp. 5505-5514, 2018.

[32] S. Iizuka, E. Simo-Serra, H. Ishikawa, "Globally and locally consistent image completion," *ACM Trans. Graph.*, vol. 36, no. 4, pp. 1-14, Aug. 2017.

[33] E. J. Candes, and Y. Plan, "Matrix completion with noise," *Proc. IEEE*, vol. 98, no. 6, pp. 925-936, Jun. 2010.

[34] I. Markovsky, "Structured low-rank approximation and its applications," *Automatica*, vol. 44, no. 4, pp. 891-909, Apr. 2008.

[35] K. H. Jin, and J. C. Ye, "Annihilating filter-based low-rank Hankel matrix approach for image inpainting," *IEEE Trans. Image Process.*, vol. 24, no. 11, pp. 3498-3511, 2015.

[36] M. Sznaier, and O. Camps, "A Hankel operator approach to texture modelling and inpainting," *Proc. Int. Workshop on Texture,* vol. 2, pp. 125-130, 2005.

[37] R. Sasaki, K. Konishi, T. Takahashi, and T. Furukawa, "Low-rank and locally linear embedding approach to image inpainting," *2018 IEEE Vis. Commun. Image Process.*, pp. 1-4, Dec. 2018.

[38] S. Razavikia, A. Amini, and S. Daei, "Reconstruction of binary shapes from blurred images via Hankel-structured low-rank matrix recovery," *IEEE Trans. Image Process.*, vol. 29, pp. 2452-2462, 2019.

[39] Y. Chen and Y. Chi, "Spectral compressed sensing via structured matrix completion," *Proc. Int. Conf. on Mach. Learn.*, no. 3, vol. 28, pp. 414-422, 2013.

[40] K. H. Jin and J. C. Ye, "Sparse and low rank decomposition of a Hankel structured matrix for impulse noise removal," *IEEE Trans. Image Process.*, vol. 27, no. 3, pp. 1448-1461, Mar. 2018.

[41] D. L. Donoho, "Compressed sensing," *IEEE Trans. on Inform. Theory*, vol. 52, no. 4, pp. 1289-1306, 2006.

[42] E. J. Candes, and Y. Plan, "Matrix completion with noise," *Procee. IEEE*, vol. 98, no. 6, pp. 925-936, Jun. 2010.

[43] S. Razavikia, A. Amini, S. Daei, "Reconstruction of binary shapes from blurred images via Hankel-structured low-rank matrix recovery," *IEEE Trans. Image Process.*, vol. 29, pp. 2452-2462, 2019.

[44] P. Kloeden and E. Platen, "Stochastic differential equations," *Numer. Solution of Stochastic Differ. Equ.,* vol. 23, pp. 103-160, 1992.

[45] B. D. Anderson, "Reverse-time diffusion equation models," *Stochastic Proc. Appl.*, vol. 12, no. 3, pp. 313-326, May. 1982.

[46] Z. Wen, W. Yin, and Y. Zhang, "Solving a low-rank factorization model for matrix completion by a nonlinear successive over-relaxation algorithm," *Math. Program. Comput.*, vol. 4, no. 4, pp. 333–361, Jul. 2012.

[47] E. J. Candès and B. Recht, "Exact matrix completion via convex optimization," *Found. of Comput. Math.*, vol. 9, no. 6, pp. 717-772, Dec. 2009.

[48] B. Recht, M. Fazel, and P. A. Parrilo, "Guaranteed minimum-rank solutions of linear matrix equations via nuclear norm minimization," *SIAM Rev.*, vol. 52, no. 3, pp. 471-501, 2010.

[49] J. Wright, A. Ganesh, S. Rao, *et al.* "Robust principal component analysis: Exact recovery of corrupted low-rank matrices via convex optimization," *Adv. Neural Inf. Process. Syst.*, pp. 2080-2088, 2009.

[50] J. F. Cai, E. J. Candès, and Z. Shen, "A singular value thresholding algorithm for matrix completion," *SIAM J. on Optim.*, vol. 20, no. 4, pp. 1956-1982, 2010.

[51] R. H. Keshavan, A. Montanari, and S. Oh, "Matrix completion from a few entries," *IEEE Trans. on Inf. Theory*, vol. 56, no. 6, pp. 2980-2998, Jun. 2010.

[52] S. Boyd, N. Parikh, E. Chu, B. Peleato, and J. Eckstein, "Distributed optimization and statistical learning via the alternating direction method of multipliers," *Found. and Trends in Mach. Learn.*, vol. 3, no. 1, pp. 1-122, Jun. 2011.

[53] Y. Song, S. Garg, J. Shi, S. Ermon, "Sliced score matching: A scalable approach to density and score estimation," *Uncertainty in Artif. Intel.*, pp. 574-584, 2020.

[54] H. Takeda, S. Farsiu, and P. Milanfar, "Kernel regression for image processing and reconstruction," *IEEE Trans. Image Process.*, vol. 16, no. 2, pp. 349-366, Feb. 2007.

[55] M. Aharon, M. Elad, and A. Bruckstein, "K-SVD: An algorithm for designing overcomplete dictionaries for sparse representation," *IEEE Trans. Signal Process.*, vol. 54, no. 11, pp. 4311-4322, Nov. 2006.

[56] J. Li, N. Wang, L. Zhang, B. Du and D. Tao, "Recurrent feature reasoning for image inpainting," *Proc. IEEE Conf. Comput. Vis. Pattern Recognit.*, pp. 7760-7768, Jun. 2020.

[57] J. Li, W. Li, *et al.* "Wavelet transform-assisted adaptive generative modeling for colorization," *IEEE Trans. Multimedia.*, pp. 1-13, 2022.